\documentclass[10pt,twocolumn,letterpaper]{article}

\usepackage{cvpr}              

\usepackage{algorithm}
\usepackage{algpseudocode}
\usepackage{amsfonts}
\usepackage{multirow}
\usepackage{amssymb} 
\usepackage{graphicx} 
\usepackage{pifont}

\usepackage{colortbl}
\definecolor{grey1}{gray}{0.85}
\definecolor{greyG}{rgb}{0.94, 0.88, 0.88}  
\definecolor{greyR}{rgb}{0.88, 0.94, 0.88}
\definecolor{greyB}{rgb}{0.96, 0.96, 0.85}

\definecolor{cvprblue}{rgb}{0.21,0.49,0.74}
\usepackage[pagebackref,breaklinks,colorlinks,allcolors=cvprblue]{hyperref}

\def\paperID{15828} 
\def\confName{CVPR}
\def\confYear{2026}

\title{Bootstrapping Video Semantic Segmentation Model \\ via Distillation-assisted Test-Time Adaptation}

\author{Jihun Kim* \\
KAIST\\
{\tt\small jihun1998@kaist.ac.kr}
\and
Hoyong Kwon*\\
KAIST\\
{\tt\small kwonhoyong3@kaist.ac.kr}
\and
Hyeokjun Kweon*\\
Chung-Ang University\\
{\tt\small hyeokjunkweon@cau.ac.kr}
\and
Kuk-Jin Yoon\\
KAIST\\
{\tt\small kjyoon@kaist.ac.kr}
}

\begin{document}
\maketitle
\def\thefootnote{*}\footnotetext{These authors contributed equally to this work.}
\begin{abstract}
Fully supervised Video Semantic Segmentation (VSS) relies heavily on densely annotated video data, limiting practical applicability. Alternatively, applying pre-trained Image Semantic Segmentation (ISS) models frame-by-frame avoids annotation costs but ignores crucial temporal coherence. Recent foundation models such as SAM2 enable high-quality mask propagation yet remain impractical for direct VSS due to limited semantic understanding and computational overhead. In this paper, we propose \textbf{DiTTA (Distillation-assisted Test-Time Adaptation)}, a novel framework that converts an ISS model into a temporally-aware VSS model through efficient test-time adaptation (TTA), without annotated videos. DiTTA distills SAM2's temporal segmentation knowledge into the ISS model during a brief, single-pass initialization phase, complemented by a lightweight temporal fusion module to aggregate cross-frame context. Crucially, DiTTA achieves robust generalization even when adapting with highly limited partial video snippets (\textit{e.g.}, initial 10\%), significantly outperforming zero-shot refinement approaches that repeatedly invoke SAM2 during inference. Extensive experiments on VSPW and Cityscapes demonstrate DiTTA’s effectiveness, achieving competitive or superior performance relative to fully-supervised VSS methods, thus providing a practical and annotation-free solution for real-world VSS tasks. The code is available at \href{https://github.com/jihun1998/DiTTA}{https://github.com/jihun1998/DiTTA}.
\end{abstract}    
\section{Introduction}
\label{sec:intro}

Video Semantic Segmentation (VSS) is a core task in understanding dynamic scenes. 
However, conventional works typically rely on large-scale annotated video datasets such as VSPW~\cite{vspw}, whose construction demands extensive pixel-level labeling and substantial human effort.

\begin{figure}[t!]
    \centering
    \includegraphics[width=0.99\linewidth]{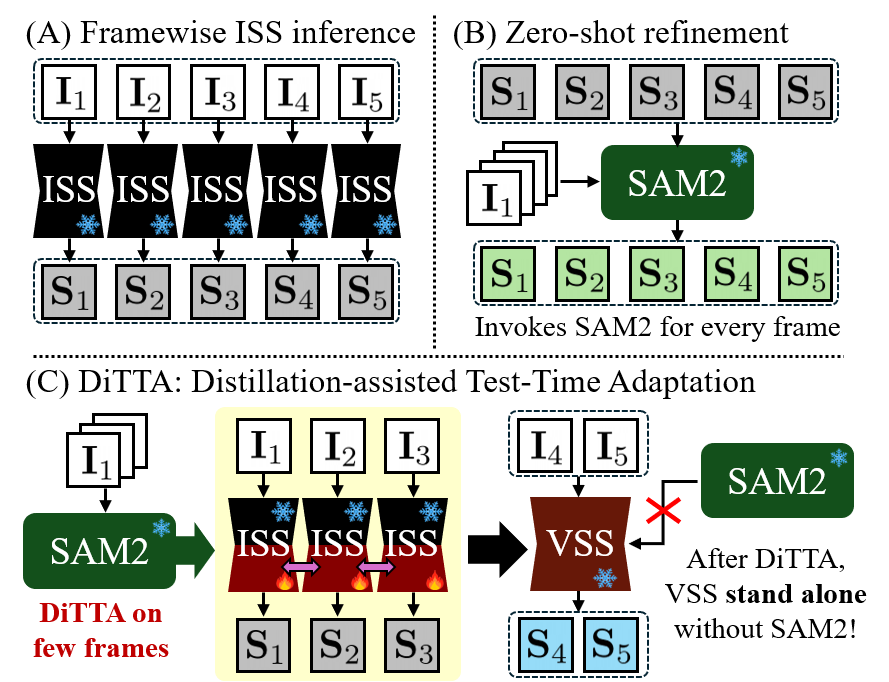}
    \vspace{-5pt}
    \caption{VSS using ISS model and SAM2. $\mathbf{I}$ and $\mathbf{S}$ are frames and their semantic segmentation results, respectively. In (C), DiTTA adapts the ISS model to the VSS model at test time by distilling temporal knowledge from SAM2 over a few initial frames. The pink arrow denotes an temporal fusion add-on.}
    \vspace{-15pt}
    \label{fig:intro}
\end{figure}

In contrast, Image Semantic Segmentation (ISS) models are trained solely on static images~\cite{maskformer, segformer, ocrnet,kwon2024phase,kweon2021unlocking,kweon2023weakly,yoon2022adversarial,yoon2024class,yoon2024diffusion}, which are significantly easier to collect and organize into large-scale annotated datasets.
Consequently, a common workaround for VSS is to apply pre-trained ISS models in a frame-by-frame manner (Fig.~\ref{fig:intro}A).
By directly employing ISS models, this strategy circumvents the need for expensive annotated video datasets.
However, it processes each frame in isolation, disregarding the temporal continuity in video. 
Moving objects typically follow coherent trajectories, which provide strong temporal cues, especially under occlusion. 
Neglecting these cues often leads to inconsistent predictions across frames.

To address this lack of temporal reasoning in frame-wise inference, we explore the use of SAM2~\cite{sam2}, a recently introduced open-source foundation model for promptable video segmentation. 
Our motivation stems from its capability for high-precision temporal mask propagation and strong generalization. 
A straightforward way to leverage this potential is to refine frame-wise ISS predictions using SAM2 in a zero-shot post-processing manner (Fig.~\ref{fig:intro}B). 
Specifically, object-level visual prompts extracted from the initial ISS output are propagated temporally using SAM2, refining the segmentation predictions on subsequent frames.

While this refinement approach leverages SAM2's temporal capabilities, it still suffers from several limitations that hinder practical deployment. 
First, it incurs substantial computational and memory overhead, as SAM2 must be invoked on every frame to track multiple objects throughout the video. 
New objects must be assigned for tracking whenever they appear, and it remains unclear when to discard obsolete ones. 
Second, this zero-shot approach relies heavily on the initial ISS results, which may be inaccurate. 
Any errors in the initial predictions directly propagate across frames, offering limited capacity for correction.

We propose a novel and adaptive strategy, named \textbf{\textit{Distillation-assisted Test-Time Adaptation (DiTTA)}}.  
DiTTA adapts a pre-trained ISS model to a given test video, effectively transforming it into a video-specific VSS model (Fig.~\ref{fig:intro}C).
Since standard ISS models operate on a single frame at a time, we devise a lightweight add-on for temporal fusion that aggregates information across frames.

The key novelty of DiTTA lies in test-time knowledge distillation from SAM2. 
Specifically, DiTTA enables the ISS model to acquire SAM2’s temporal segmentation capability by mimicking its spatio-temporal mask predictions.
Importantly, this distillation is performed \textbf{using only the initial few frames} of the test video—without access to the full video.  
During this, the ISS model learns SAM2’s temporal segmentation behavior by mimicking its outputs, and is bootstrapped into a temporally-aware VSS model tailored to the input video.
Once adaptation is completed, the resulting model is used for inference on subsequent frames without any further adaptation or reliance on SAM2.  
In contrast to computationally intensive zero-shot refinement that invoke SAM2 repeatedly at inference time, DiTTA enables fast and consistent video segmentation.

We validate our method through comprehensive experiments, comparing against three representative baselines: (1) standard ISS/VSS models, (2) existing TTA methods, and (3) zero-shot refinement with SAM2. 
We further demonstrate strong generalization to domain-shifted scenarios and different training data regimes, as well as competitive performance under full-video adaptation.
These results highlight DiTTA as a practical, data-efficient, and scalable solution for real-world video segmentation tasks.

\section{Related Works}
\label{sec:rw}

\subsection{Video Semantic Segmentation (VSS)}
VSS~\cite{netwarp, vpseg, accel_cvpr19, pearl_iccv17, gsvnet_icme21, stt_icmm21, low_cvpr18, etc_eccv20, vspw, stgru_cvpr18, motion_cvprw23, cffm, mrcfa_eccv22, tma_icip21,cheng2022masked,yang2024end,cho2024finding} aims to predict pixel-wise classes for each frame in a video. 
Previous VSS studies have primarily focused on leveraging temporal information between frames to exploit the sequential nature of video data. 
One widely used approach is to incorporate optical flow~\cite{flownet_iccv15} between sequential frames. 
Methods following this approach~\cite{joint_aaai20, netwarp, rta_mc_eccv18, accel_cvpr19, stgru_cvpr18} have devised additional modules that estimate optical flow and integrate it at the feature level to capture temporal dynamics. 
Recently, MPVSS~\cite{mpvss_nips23} introduces a query-based flow method to propagate masks from keyframes, improving efficiency.
Another approach leverages attention mechanisms to capture temporal dependencies across frames~\cite{tdnet,stt_icmm21,lmanet, motion_cvprw23,cffm, mrcfa_eccv22, tma_icip21}. 
For example, TDNet~\cite{tdnet} employs an attention mechanism within a temporally distributed network, enabling faster inference. 
CFFM~\cite{cffm} proposes coarse-to-fine feature mining modules, while MRCFA~\cite{mrcfa_eccv22} introduces cross-frame affinity to make effective use of temporal information. 
VPseg~\cite{vpseg}, specifically designed for driving scenes, incorporates the concept of a vanishing point as a valuable prior for this setting.
Despite significant progress in VSS, we find that conventional VSS studies have been heavily dependent on specialized VSS datasets, which limits their practicality in real-world applications. 

\subsection{Using External Source for Video Segmentation}
Extensive research on video segmentation has leveraged external sources, which can generally be categorized into two approaches: (1) boundary estimation models (\textit{e.g.}, superpixel) designed to effectively capture segment structures \cite{vss_sp2,vss_sp1,vss_sp4,vss_sp3}, and (2) temporal knowledge sources, such as optical flow, that provide insights into motion context between frames~\cite{netwarp, rta_mc_eccv18,accel_cvpr19, stgru_cvpr18,huang2022minvis,lo2023spatio}.
Additionally, recent studies have explored segmentation using segmentation knowledge from SAM/SAM2 \cite{sam1,sam2}. This includes applications not only in general video segmentation \cite{sam2chal2,sam2chal3,sam2chal1, ye2025entitysam} but also in specialized domains such as camouflaged object detection~\cite{sam2camo2,sam2camo1} and medical imaging~\cite{sam2med2,sam2med1}, where SAM-based segmentation has shown promising results~\cite{kweon2024sam,kweon2025wish}.
However, these approaches differ significantly from our method in their specific use of SAM2, typically as a post-processing step for pre-trained video segmentation models or by fine-tuning SAM2 with a specific video segmentation dataset.
In contrast, we avoid the use of any VSS dataset, thereby alleviating the substantial burden of dataset construction, which is often a major challenge in practice. 
To address this, the proposed DiTTA adapts an ISS model to video at test-time, leveraging SAM2 as a source of video knowledge.

\subsection{Test-Time Adaptation (TTA)}
TTA~\cite{adacontrast,colomer2023adapt,fleuret2021uncertainty, shot_pmlr20, sar_iclr23, auxadapt_cvpr, tesla_cvpr23, volpi2022road, wang2020tent, cotta, wang2023dynamically, DSS_wacv24, auxadapt_wacv} adapts a pre-trained model to a target test domain—typically a small set of test data—in an unsupervised manner without access to source domain data. 
In TTA research, various online optimization strategies have been investigated, including entropy minimization~\cite{fleuret2021uncertainty, shot_pmlr20, sar_iclr23, wang2020tent}, pseudo-labeling~\cite{fleuret2021uncertainty, shot_pmlr20, tesla_cvpr23, DSS_wacv24,jang2024talos,kim2025dc}, and contrastive learning~\cite{adacontrast}.
However, these methods predominantly target classification tasks, with a few recent studies~\cite{colomer2023adapt,volpi2022road,wang2023dynamically} exploring TTA for semantic segmentation in static images.
Research specifically addressing TTA in the context of VSS remains exceedingly limited, making our approach to adapting ISS models to the video domain a significant and novel contribution. 
To the best of our knowledge, the only similar work is AuxAdapt~\cite{auxadapt_cvpr,auxadapt_wacv}, which also aims to adapt ISS models for VSS through a frame-wise pseudo-labeling strategy.
However, this method falls short in fully adapting to the video domain as it overlooks temporal information across frames.
To address this limitation, we introduce a TTA approach that goes beyond self-reliant adaptation by leveraging the temporal segmentation capability of SAM2.
DiTTA performs genuine video-level adaptation, as opposed to prior frame-by-frame strategies, by explicitly incorporating temporal coherence into the adaptation.

\section{Methods}\label{sec:method}

The proposed DiTTA framework performs TTA of a pre-trained ISS model to a given video by leveraging both internal cues from the ISS model itself and distillation signals from SAM2.
DiTTA comprises three key components: (1) a lightweight attention-based add-on module that structurally enables the model to capture temporal cues across frames;   
(2) a distillation target from SAM2 to provide assistant guidance for more effective TTA; and (3) a contrastive learning objective that promotes feature alignment during TTA by enforcing object-level temporal consistency.



\subsection{Temporal Fusion via Attention-based Add-on}\label{sec:addon}

The pre-trained ISS model is inherently limited to frame-wise inference, as it operates on a single image at a time and lacks any mechanism to aggregate temporal information across frames.
To address this, we introduce a lightweight attention-based add-on module that enables the pre-trained model to utilize contextual temporal cues from adjacent frames during the TTA process.

\begin{figure*}[t]
    \centering
    \includegraphics[width=0.99\linewidth]{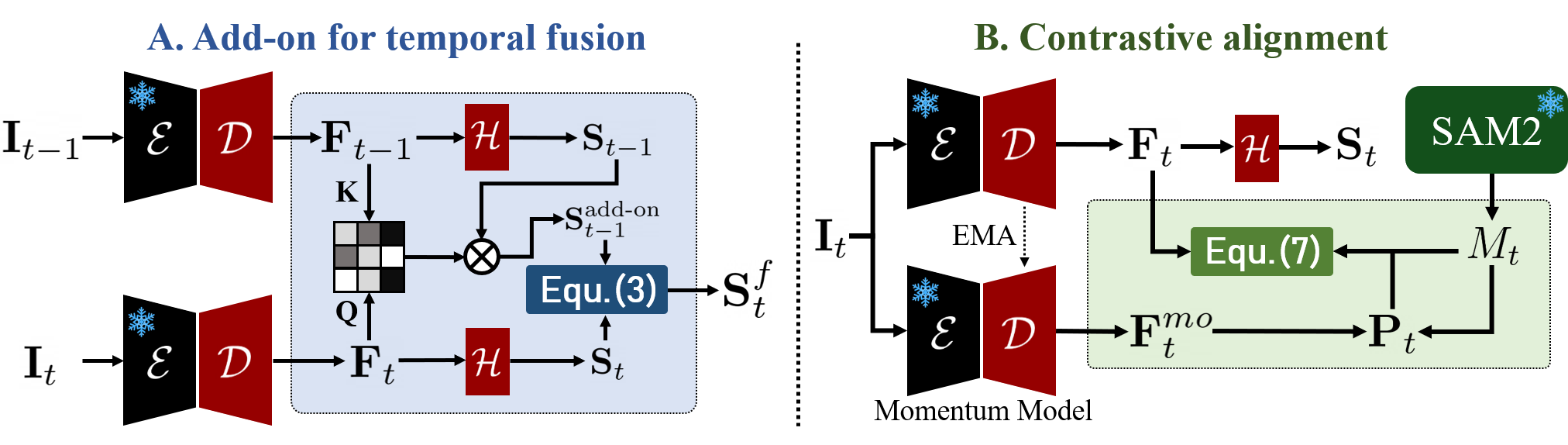}
    \vspace{-7pt}
    \caption{Overview of our DiTTA (Distillation-assisted Test-Time Adaptation) framework. It comprises a lightweight add-on for temporal fusion (left blue box) and a mask-based contrastive alignment (right green block).}
    \label{fig:framework}
    \vspace{-15pt}
\end{figure*}

As illustrated in Fig.~\ref{fig:framework}A, the add-on acts as a temporal bridge between consecutive frames, allowing the ISS model to share and integrate information across time.  
As a result, the model is effectively transformed from a purely image-based segmenter into a temporally-aware VSS model.

The add-on employs a cross-attention mechanism to integrate information from the previous frame.  
Given a pair of consecutive frames $(\mathbf{I}_{t-1}, \mathbf{I}_t)$, the ISS model produces features $(\mathbf{F}_{t-1}, \mathbf{F}_t)$ and logits $(\mathbf{S}_{t-1}, \mathbf{S}_t)$.  
These features are projected to form a query-key pair, where $\mathbf{Q}_t = \text{Proj}_Q(\mathbf{F}_t)$ and $\mathbf{K}_{t-1} = \text{Proj}_K(\mathbf{F}_{t-1})$, while $\mathbf{S}_{t-1}$ serves as the value.  
The resulting attention output is computed as
\begin{equation}\label{eq:temporal_attention}
    \mathbf{S}^\text{add-on}_{t-1} = \text{softmax}(\mathbf{Q}_t \mathbf{K}^\mathcal{T}_{t-1}) \mathbf{S}_{t-1},
\end{equation}
where $\mathbf{S}^\text{add-on}_{t-1}$ is spatially aligned to the current frame $\mathbf{I}_t$.

Once both the original and attention-informed logits are obtained, we determine how to combine them for each pixel.  
Because different frames may exhibit varying levels of confidence—due to occlusion or motion blur—we use reliability score to guide this fusion.  
Specifically, we define a pixel-wise reliability map $\mathbf{R}_t$ for each frame $\mathbf{I}_t$, based on the normalized entropy of the prediction logits $\mathbf{S}_t$:
\begin{equation}\label{eq:conf_map}
    \mathbf{R}_t(x,y) = 1 - \frac{\mathbf{E}_t(x,y)}{\max_{(x,y)} \mathbf{E}_t(x,y)},
\end{equation}
where $\mathbf{E}_t(x,y)$ denotes the pixel-wise entropy over the class distribution from $\mathbf{S}_t(x,y)$.
The final prediction $\mathbf{S}^f_t$ is then computed in a reliability-aware manner as follows:
\begin{equation}\label{eq:logit_mix}
\mathbf{S}^f_t = 
\begin{cases} 
      \mathbf{S}_t & \text{if } \mathbf{R}_t \geq \tau, \\
      \frac{\mathbf{R}_t}{\mathbf{R}_t + \mathbf{R}_{t-1}^{\text{add-on}}}\mathbf{S}_t +\frac{\mathbf{R}_{t-1}^{\text{add-on}}}{\mathbf{R}_t + \mathbf{R}_{t-1}^{\text{add-on}}}\mathbf{S}^\text{add-on}_{t-1}  & \text{otherwise},
   \end{cases}
\end{equation}
where $\tau$ is a threshold hyperparameter and the fusion is applied in a pixel-wise manner.

\subsection{Distillation Targets of DiTTA}\label{sec:pgt}

One of the fundamental challenges in TTA is the absence of supervision at test time.  
Existing methods often rely on heuristic objectives such as entropy minimization~\cite{wang2020tent}, or adopt self-supervised schemes by reusing model predictions as pseudo-labels~\cite{cotta}.  
However, since these approaches mainly depend on the model’s own predictions, the extent of improvement is often limited.

To address this, DiTTA constructs distillation targets by combining the complementary strengths of the ISS model and SAM2.  
Specifically, semantic predictions from the ISS model are used to select reliable prompt locations and to assign class labels, while SAM2 contributes temporal propagation to generate object-consistent segmentation masks.  
This division of roles allows DiTTA to harness the semantic grounding of the ISS model and the spatial-temporal consistency of SAM2 in a unified manner.  
The resulting targets serve as intermediate supervision signals, guiding both feature-level and logit-level TTA of the ISS model.

We begin by sampling high-reliability pixels from the ISS model’s predictions, using class-wise filtering and entropy-based thresholding to ensure semantic and spatial diversity.  
Given the sampled prompt set from the initial frame, we apply SAM2 to generate sets of object-consistent masks through bidirectional propagation, denoted as $\{M^i_t\}$.
While these masks do not aim to cover the entire frame, they focus on regions where the ISS model and SAM2 are confident.  
Hence, they offer a reliable guide that captures object-level temporal coherence, informed by SAM2’s segmentation capability.  
Details of this process are provided in the \textit{Supplementary Material}.

Each spatiotemporal mask $M^i = \{m^i_1, m^i_2, \dots, m^i_T\}$ is then assigned a class label $c^i \in \mathbb{C}$ using a soft scoring that aggregates per-frame ISS predictions within the mask.
To ensure robustness, the score $\alpha^c$ of each class $c$ is defined as:
\begin{equation}\label{eq:score_all}
    \alpha^c = \alpha^c_{\text{rel}} \cdot (\gamma^c_{\text{area}})^{\lambda_{\text{area}}} \cdot (\gamma^c_{\text{freq}})^{\lambda_{\text{freq}}},
\end{equation}
where $\lambda_{\text{area}}$ and $\lambda_{\text{freq}}$ are weighting hyperparameters.
Here, $\alpha^c_{\text{rel}}$ measures the average reliability of the ISS predictions for class $c$ within the mask, reflecting how confidently the model assigns that class in relevant regions.  
$\gamma^c_{\text{area}}$ denotes the fraction of the mask area predicted as class $c$, which helps to suppress small, noisy predictions.  
Finally, $\gamma^c_{\text{freq}}$ adjusts for class imbalance by favoring classes that appear less frequently across the video.
Intuitively, this favors classes that are both confidently predicted and consistently occupy a significant portion of the mask, while also compensating for long-tailed class distributions.
More details are in \textit{Supp}.

Once class labels $c^i$ are assigned to $M^i$, we apply a logit-level cross-entropy loss within the corresponding masked regions only.  
For each object $i$ and each frame $t$, the loss is computed over the region defined by the binary mask $m^i_t$:
\begin{equation}
    \mathcal{L}^{\text{Distill}}_t\! \!   = \! \!  \sum_{i} \!\!   \sum_{(x,y) \in m^i_t}\! \! \!\! \! \!\text{CE}\! \left(\mathbf{S}_t(x,y), c^i\right) \! +  \text{CE}\! \left(\mathbf{S}^{\text{add-on}}_{t-1}\! (x,y), c^i\right)\!, \!
\end{equation}
where $\text{CE}(\cdot)$ denotes the standard cross-entropy loss.
The summation is restricted to pixels within the binary mask $m^i_t$, \textit{i.e.}, regions where the distillation target is defined. 
These correspond to spatiotemporal regions determined by SAM2, while the semantic information is provided by the ISS model.

\begin{figure*}[t]
    \centering
    \includegraphics[width=0.9\linewidth]{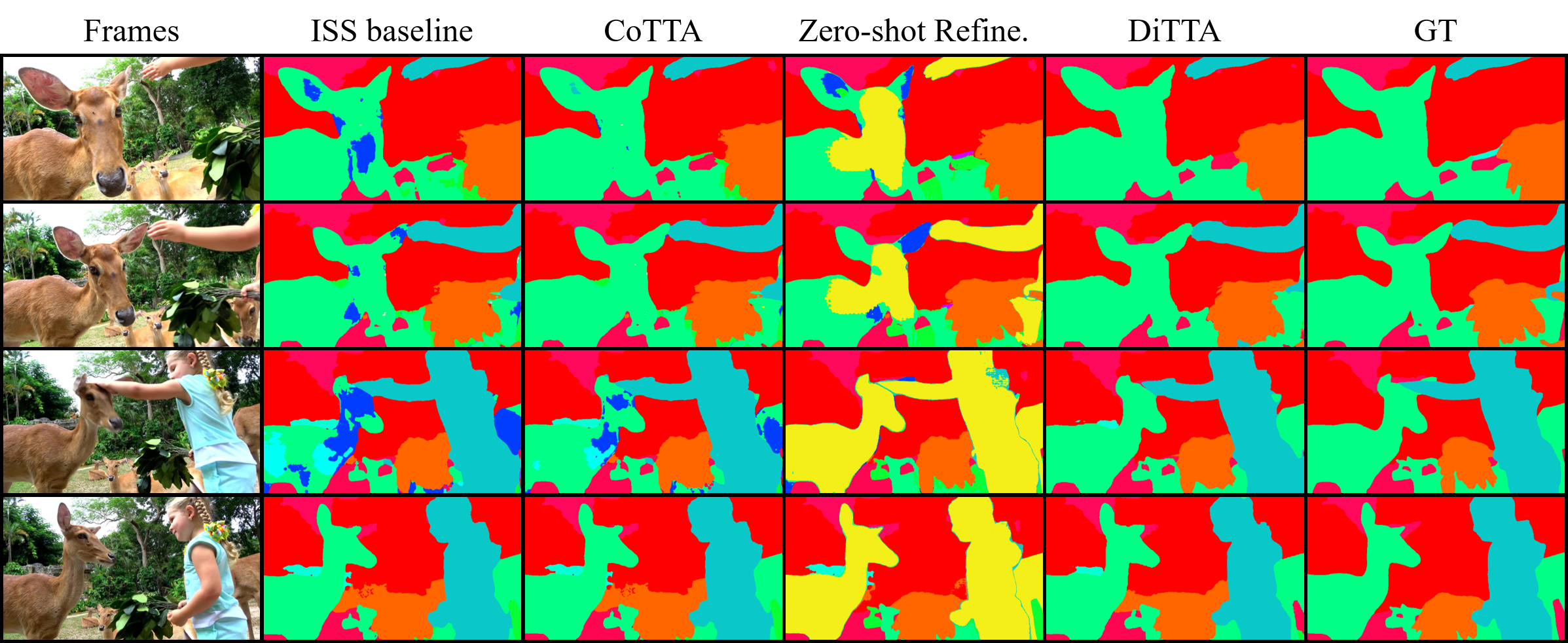}
    \vspace{-6pt}
    \caption{Qualitative comparison of VSS results across various methods under W2F protocol. Only the initial 10\% of frames are used for distillation, with evaluation on the remaining. The frames shown are entirely unseen.}
    \label{fig:qual_s2}
    \vspace{-13pt}
\end{figure*}

\subsection{Mask-based Contrastive Alignment}\label{sec:contra}
While the logit-level loss ensures semantic alignment in object regions, it does not constrain feature-level coherence across frames. 
To address this, we apply a mask-based contrastive loss~\cite{maskcontra,adacontrast} that encourages the model to produce consistent representations for the same object over time.

As depicted in Fig.~\ref{fig:framework}B, we adopt a momentum encoder framework, where object-wise prototypes are computed from the momentum branch.  
For each spatiotemporal object mask $M^i = \{m^i_1, \dots, m^i_T\}$, we compute the prototype vector $\mathbf{P}^i_t$ as:
\begin{equation}\label{eq:proto}
    \mathbf{P}_t^i = \frac{1}{|m^i_{1:t}|} \sum_{u \in [1,t]}  \sum_{(x,y) \in m_u^i} \mathbf{F}^{mo}_u(x,y) \cdot \mathbf{R}^{mo}_u(x,y) ,
\end{equation}

where $\mathbf{F}^{mo}$ is the momentum feature map and $\mathbf{R}^{mo}$ is the corresponding reliability map (Equ.~\ref{eq:conf_map}).
Given the prototype $\mathbf{P}^i_t$, the main model aligns its current features $\mathbf{F}_t$ within the same mask $M^i$ as:
\begin{equation}\label{eq:loss_contra}
    \mathcal{L}^\text{Contra}_t = -\sum_{i} \! \sum_{(x,y) \in m^i_t} \log \frac{\exp(\mathbf{F}_t(x,y) \cdot \mathbf{P}^i_t)}{\sum_{j} \exp(\mathbf{F}_t(x,y) \cdot \mathbf{P}^j_t)},
\end{equation}
where $\mathbf{F}_t(x, y)$ is the feature vector at pixel $(x, y)$, and the summation is taken over all pixels in $m^i_t$ where the mask equals 1.
The momentum encoder is updated via exponential moving average (EMA), providing stable prototype targets throughout adaptation.  
This contrastive alignment complements the logit-level loss by reinforcing object-level temporal consistency in the representation space.

DiTTA combines both logit-level and feature-level contrastive formulation derived from the distillation targets.  
To sum up, the total loss on $t$-th frame is defined as:
\begin{equation}
    \mathcal{L}^{\text{DiTTA}}_t = \mathcal{L}^{\text{Distill}}_t + \mathcal{L}^{\text{Contra}}_t.
\end{equation}

This unified objective enables the ISS model to learn temporally consistent representations for VSS through feature alignment.
Unlike zero-shot refinement, our method requires only a single pass of SAM2 during adaptation and performs efficient inference.

\section{Experimental Settings}

\subsection{Dataset}
VSPW~\cite{vspw} is the most widely adopted VSS benchmark, which contains 198,244/24,502/28,887 frames (2,806/343/387~clips) for train/val/test splits, with GT segmentation map across 124 categories.
Notably, VSPW is annotated at 15fps, in contrast to other VSS datasets like Cityscapes~\cite{cityscapes}, which include a single annotated frame per clip. 
This limited availability—practically restricted to VSPW—highlights the value of our approach.

\subsection{Evaluation Protocol}

Leveraging open-source vision foundation models such as SAM2~\cite{sam2} holds great promise for VSS. However, refining segmentation results—or adapting the models—on every frame of a video incurs significant computational and memory overhead, making it impractical for many deployment scenarios. This limitation can become especially critical in resource-constrained environments such as on-robot inference or surveillance.

To better reflect such constraints, we introduce a realistic evaluation setting, which we term \textbf{Warm-Up then Freeze (W2F)}. In the W2F protocol, test-time adaptation is performed only on an initial portion of the video (as a warm-up), after which the adapted model is frozen and used solely for inference on the remaining frames. 

For example, with a 10\% warm-up ratio, only the first 10\% of frames are used for adaptation; the resulting model is then evaluated on the remaining 90\% without further updates.
This setting allows us to assess how effectively a model can leverage temporal knowledge acquired from a short initial segment, while adhering to practical deployment constraints. 
Unless otherwise specified, all experiments are conducted under the W2F protocol.

To evaluate semantic segmentation performance, we use mean IoU (mIoU) as primary metrics, following~\cite{netwarp, cffm, mrcfa_eccv22}. 
Additionally, we employ weighted IoU (wIoU) and mean video consistency (mVC)~\cite{vspw}. mVC is for assessing the smoothness of the predicted results across frames.

\begin{table*}
\centering
\vspace{-5pt}
\caption{Performance comparison of ISS, VSS, ISS+TTA, zero-shot refinement, and DiTTA under the W2F protocol.
DiTTA does not involve SAM2 inference during evaluation, unlike zero-shot refinement.}
\vspace{-5pt}
\resizebox{0.95\linewidth}{!}{
\begin{tabular}{c|c|c|c|c|c|c|c}
\hline
\textbf{Ratio} & \textbf{Approach} & \textbf{Methods} & \textbf{FPS} $\uparrow$ & \textbf{mIoU} $\uparrow$ & \textbf{wIoU} $\uparrow$ &  \textbf{mVC$_8$} $\uparrow$ & \textbf{mVC$_{16}$} $\uparrow$  \\ \hline\hline

\multirow{8}{*}{10\%} & ISS & \cellcolor{greyR}SegFormer~\cite{segformer} & \cellcolor{greyR}18.58 & \cellcolor{greyR}49.0 & \cellcolor{greyR}66.3 & \cellcolor{greyR}88.3 & \cellcolor{greyR}84.3 \\ \cline{2-8}

& \multirow{3}{*}{VSS} & CFFM~\cite{cffm} & 5.98 & 49.4 & 66.5 & 90.8 & 87.3  \\ 
& & CFFM++~\cite{cffmpp} & 5.85 & 49.6 & 66.1 & 90.4 & 86.8 \\ 
& & TV3S~\cite{hesham2025exploiting} & 5.71 & 49.8 & 66.4 & 91.5 & 88.0 \\ \cline{2-8}

& \multirow{3}{*}{ISS+TTA} & TENT~\cite{wang2020tent} & 18.45 & 49.6 & 66.5 & 90.3 & 87.2 \\ 
& & AuxAdapt~\cite{auxadapt_cvpr} & 18.56 & 48.7 & 65.5 & 89.4 & 85.9 \\ 
& & CoTTA~\cite{cotta} & 18.48 & 49.6 & 66.7 & 89.7 & 86.4 \\ \cline{2-8}

& \multirow{2}{*}{ISS+SAM2} & Zero-shot Refine.  & 1.41 & 49.7 & 66.5 & 94.7 & 92.9 \\ 
& & \cellcolor{greyR}DiTTA (Ours) & \cellcolor{greyR}13.45 & \cellcolor{greyR}51.1 \textcolor{blue}{(+2.1)} & \cellcolor{greyR}66.5 \textcolor{blue}{(+0.2)} & \cellcolor{greyR}94.1 \textcolor{blue}{(+5.8)} &\cellcolor{greyR} 92.2 \textcolor{blue}{(+7.9)} \\ \hline\hline
\multirow{8}{*}{25\%} &          ISS         & \cellcolor{greyG}SegFormer~\cite{segformer} &\cellcolor{greyG} 18.58 &\cellcolor{greyG} 49.0 & \cellcolor{greyG}66.4 & \cellcolor{greyG}88.3 & \cellcolor{greyG}84.3 \\ \cline{2-8}

& \multirow{3}{*}{VSS} & CFFM~\cite{cffm} & 5.98 & 49.5 & 66.5 & 90.8 & 87.3 \\ 
&                      & CFFM++~\cite{cffmpp} & 5.85 & 49.7 & 66.1 & 90.4 & 86.8 \\ 
& & TV3S~\cite{hesham2025exploiting} & 5.71 & 49.9 & 66.6 & 91.5 & 88.0 \\ \cline{2-8}

& \multirow{3}{*}{ISS+TTA} & TENT~\cite{wang2020tent} & 18.45 & 48.7 & 65.6 & 91.6 & 88.9 \\ 
&                          & AuxAdapt~\cite{auxadapt_cvpr} & 18.56 & 48.7 & 65.6 & 90.2 & 86.9 \\ 
&                          & CoTTA~\cite{cotta} & 18.48  & 49.8 & 66.8 & 90.3 & 87.1 \\ \cline{2-8}

& \multirow{2}{*}{ISS+SAM2} & Zero-shot Refine.  & 1.41  & 49.5 & 66.4 & 94.8 & 93.1\\ 
& & \cellcolor{greyG}DiTTA (Ours) & \cellcolor{greyG}13.45 & \cellcolor{greyG}51.0 \textcolor{blue}{(+2.0)} & \cellcolor{greyG}66.6 \textcolor{blue}{(+0.2)} & \cellcolor{greyG}94.5 \textcolor{blue}{(+6.2)} & \cellcolor{greyG}92.6 \textcolor{blue}{(+8.3)} \\ \hline\hline
\multirow{8}{*}{50\%} & ISS & \cellcolor{greyB}SegFormer~\cite{segformer} & \cellcolor{greyB}18.58 & \cellcolor{greyB}48.7 &\cellcolor{greyB} 66.4 &\cellcolor{greyB} 88.3 &\cellcolor{greyB} 84.2 \\ \cline{2-8}

& \multirow{3}{*}{VSS} & CFFM~\cite{cffm} & 5.98 & 49.7 & 66.5 & 90.8 & 87.2 \\ 
&                      & CFFM++~\cite{cffmpp} & 5.85 & 49.7 & 66.2 & 90.6 & 86.9  \\ 
& & TV3S~\cite{hesham2025exploiting} & 5.71 & 49.8 & 66.7 & 91.3 & 87.6 \\ \cline{2-8}

& \multirow{3}{*}{ISS+TTA} & TENT~\cite{wang2020tent} & 18.45 & 47.6 & 64.3 & 92.6 & 90.3 \\ 
&                          & AuxAdapt~\cite{auxadapt_cvpr} & 18.56 &48.7 & 65.6 & 90.5 & 87.3  \\ 
&                          & CoTTA~\cite{cotta} & 18.48 & 49.8 & 66.7 & 90.6 & 87.5 \\ \cline{2-8}

& \multirow{2}{*}{ISS+SAM2} & Zero-shot Refine.  & 1.41 & 50.1 & 67.3 & 95.0 & 93.3  \\ 
& & \cellcolor{greyB}DiTTA (Ours) & \cellcolor{greyB}13.45 & \cellcolor{greyB}52.3 \textcolor{blue}{(+3.6)} &\cellcolor{greyB} 67.1 \textcolor{blue}{(+0.7)} & \cellcolor{greyB}94.9 \textcolor{blue}{(+6.6)} & \cellcolor{greyB}93.0 \textcolor{blue}{(+8.8)} \\ \hline

\end{tabular}
\label{tab:scenario2}
}
\vspace{-8pt}
\end{table*}

\subsection{Implementation Details}
We employ SegFormer~\cite{segformer} with a MiT-B5 backbone, pre-trained on the VSPW train set, as the default ISS model.
This model is trained in a frame-wise manner, not with video.
For TTA, only the parameters in the decoder of ISS model are updated. 
We set the learning rate to 0.001, with 5 iterations per frame. 
For hyperparameters, we set $\tau=0.8$, $\lambda_{\text{area}}=0.3$, and $\lambda_{\text{freq}}=0.8$.
All experiments are conducted with a fixed seed.
More details are explained in \textit{Supplementary Material}.

\section{Main Results}\label{sec:exp}

Table~\ref{tab:scenario2} shows a comprehensive comparison between DiTTA and existing approaches under the W2F protocol.  
To the best of our knowledge, no prior work follows the exact same setting as ours. 
Therefore, we organize our comparisons into three representative categories: (1) direct application of ISS/VSS models, (2) ISS combined with TTA methods, and (3) ISS combined with zero-shot refinement. 
All comparisons are conducted under three different warm-up ratios: 10\%, 25\%, and 50\%.
Figure~\ref{fig:qual_s2} shows qualitative comparisons and more are in \textit{Supplementary Material}.

\subsection{Comparison with ISS/VSS}
The first group of baselines includes standard ISS and VSS models trained on the VSPW dataset without any form of test-time adaptation. 
For consistency, we fix the backbone architecture to SegFormer~\cite{segformer}, a representative ISS model. As VSS baselines, we include CFFM~\cite{cffm}, CFFM++~\cite{cffmpp} and TV3S~\cite{hesham2025exploiting}, which use the same backbone and are trained directly on video data.
The ISS model is the baseline corresponds to the pre-trained model used in DiTTA. 

As shown in Table~\ref{tab:scenario2}, DiTTA consistently outperforms both ISS and VSS baselines across all warm-up ratios.
Remarkably, despite being adapted from an ISS model, DiTTA even surpasses fully supervised VSS models.
For example, even with only 10\% of the video used for adaptation, DiTTA improves mIoU by +2.1\%p over ISS baseline and +1.5\%p over the CFFM++.
Considering that the VSPW dataset~\cite{vspw} contains over 120 semantic categories, such performance gains demonstrate strong gain in VSS capability.

Since FPS is also a crucial factor in VSS, we measure the inference speed of each method on a single GeForce RTX 3090 GPU for a fair comparison.
DiTTA runs significantly faster than VSS models while achieving superior segmentation performance, demonstrating its effectiveness in both VSS capability and efficiency.
Compared to the original ISS model, DiTTA exhibits a modest reduction in FPS due to the lightweight temporal fusion module; however, this minor overhead is well justified by the substantial performance improvement it brings.
ISS-based TTA methods perform nearly identical to the original ISS in terms of runtime, yet offer relatively minor accuracy gains.
Overall, these results indicate that DiTTA, even when adapted using a limited number of frames, can substantially enhance both spatial accuracy and temporal consistency over ISS models.

\subsection{Comparison with TTA}
We also compare DiTTA against existing TTA methods for ISS.
These methods adapt the model to test-time video in a frame-wise fashion, without explicit temporal modeling.
For fair comparison, we apply TENT~\cite{wang2020tent}, AuxAdapt~\cite{auxadapt_cvpr}, and CoTTA~\cite{cotta} to the same SegFormer backbone.

Table~\ref{tab:scenario2} shows that DiTTA consistently outperforms all TTA baselines across all warm-up ratios.
For example, with a 10\% warm-up ratio, DiTTA achieves a +1.5\%p mIoU improvement over the strongest baseline (CoTTA).
These results suggest that transferring structured temporal knowledge from SAM2 via distillation is more effective than relying solely on unsupervised adaptation objectives.
In terms of efficiency, DiTTA introduces an add-on for temporal adaptation, resulting in a modest reduction in FPS compared to pure ISS+TTA methods.
Nonetheless, it maintains efficiency at over 13 FPS, while offering significantly higher segmentation quality.
This trade-off between accuracy and efficiency highlights DiTTA as a practical and scalable solution for real-world deployment scenarios.

\subsection{Comparison with Zero-shot Refinement}

A natural question arising from the previous comparisons is whether DiTTA’s improvements simply result from the use of SAM2.
To clarify this, we compare DiTTA against the zero-shot refinement strategy, which also utilizes SAM2 but without any model adaptation.
In this zero-shot refinement approach, per-class instance masks are extracted from the ISS output on the first frame using 8-connected component labeling.
Masks occupying less than 0.05\% of the image area are discarded, and each remaining mask is assigned a confidence score equal to the maximum ISS confidence of its constituent pixels.
These masks are then propagated to subsequent frames using SAM2, and for each frame, the propagated masks are merged with the ISS predictions.

\begin{table}[t]
\caption{Ablation studies of DiTTA's three components. Experiments are conducted under the 50\% W2F protocol.}
\centering
\vspace{-7pt}
\resizebox{0.99\linewidth}{!}{
\begin{tabular}{c|c|c|c|c}
\hline
\textbf{Exp.} & \textbf{Add-on} & \textbf{Distill. Targets}     & \textbf{Contrast.}                                                                                 & \textbf{mIoU} $\uparrow$ \\ 
\hline\hline
ISS          &                          &                                &                           &        48.7               \\ \hline
A          &      \checkmark                     &                                &                           &        49.9               \\ \hline
B          &                          &                 \checkmark                &                           &        50.8          \\ \hline
C          &                          &                                &                  \checkmark          &        50.2              \\ \hline
\rowcolor{grey1} DiTTA             & \checkmark & \checkmark  & \checkmark & 52.3                      \\ \hline
\end{tabular}
}
\label{tab:abl}
\vspace{-17pt}
\end{table}

Compared to this strategy, \textbf{DiTTA offers both higher efficiency and stronger performance}.
It eliminates the need to invoke SAM2 repeatedly during inference, while consistently outperforming zero-shot refinement across all warm-up ratios.
For instance, with only a 10\% warm-up ratio, DiTTA achieves an mIoU improvement of +1.4\%p and operates almost ten times faster (13.45 FPS vs. 1.41 FPS).

The performance gap becomes even larger at higher warm-up ratios, suggesting that DiTTA effectively transfers SAM2’s temporal knowledge into the ISS model during adaptation.
These results suggest that DiTTA can serve as a more effective and efficient alternative to the zero-shot refinement relying on repeated use of SAM2.

\section{Additional Analysis and Comparisons}

\subsection{Ablation Studies}
We conduct ablation studies to show the contributions of DiTTA’s key components: the temporal fusion module, SAM2-based distillation targets, and contrastive alignment (Table~\ref{tab:abl}).
In the variant without distillation targets, we follow standard practice in prior TTA literature~\cite{auxadapt_cvpr,cotta,auxadapt_wacv} by using the ISS baseline predictions as self-supervised targets.

The results show that each component contributes meaningful performance gains over the ISS baseline (48.7 mIoU).
Exp A leverages temporal context, Exp B introduces external guidance, and Exp C improves feature alignment—each contributing to performance gains.
When all three components are combined, DiTTA achieves the highest performance (52.3 mIoU), demonstrating their complementary effects.
These validates the design of DiTTA as a unified and modular framework.

\begin{table}[t]
\caption{Comparison on Cross-dataset setting. Methods are trained on VSPW~\cite{vspw} and evaluated on Cityscapes~\cite{cityscapes}.}
\vspace{-6pt}
\centering
\resizebox{0.99\linewidth}{!}{
\begin{tabular}{r|c|c|c|c}
\hline
\multicolumn{1}{c|}{}  & ISS & CFFM  & CoTTA & \cellcolor{grey1}DiTTA (Ours) \\ \hline
\textbf{mIoU }$\uparrow$                  & 44.2     & 46.1  & 46.3 & \cellcolor{grey1}46.9 \textcolor{blue}{(+2.7)}         \\ \hline
\textbf{wIoU}    $\uparrow$               & 74.2     & 76.4  & 76.9 & \cellcolor{grey1}77.9 \textcolor{blue}{(+3.7)}         \\ \hline
\end{tabular}
}
\vspace{-5pt}
\label{tab:quan_city}
\end{table}

\begin{figure}[t!]
    \centering
    \includegraphics[width=0.99\linewidth]{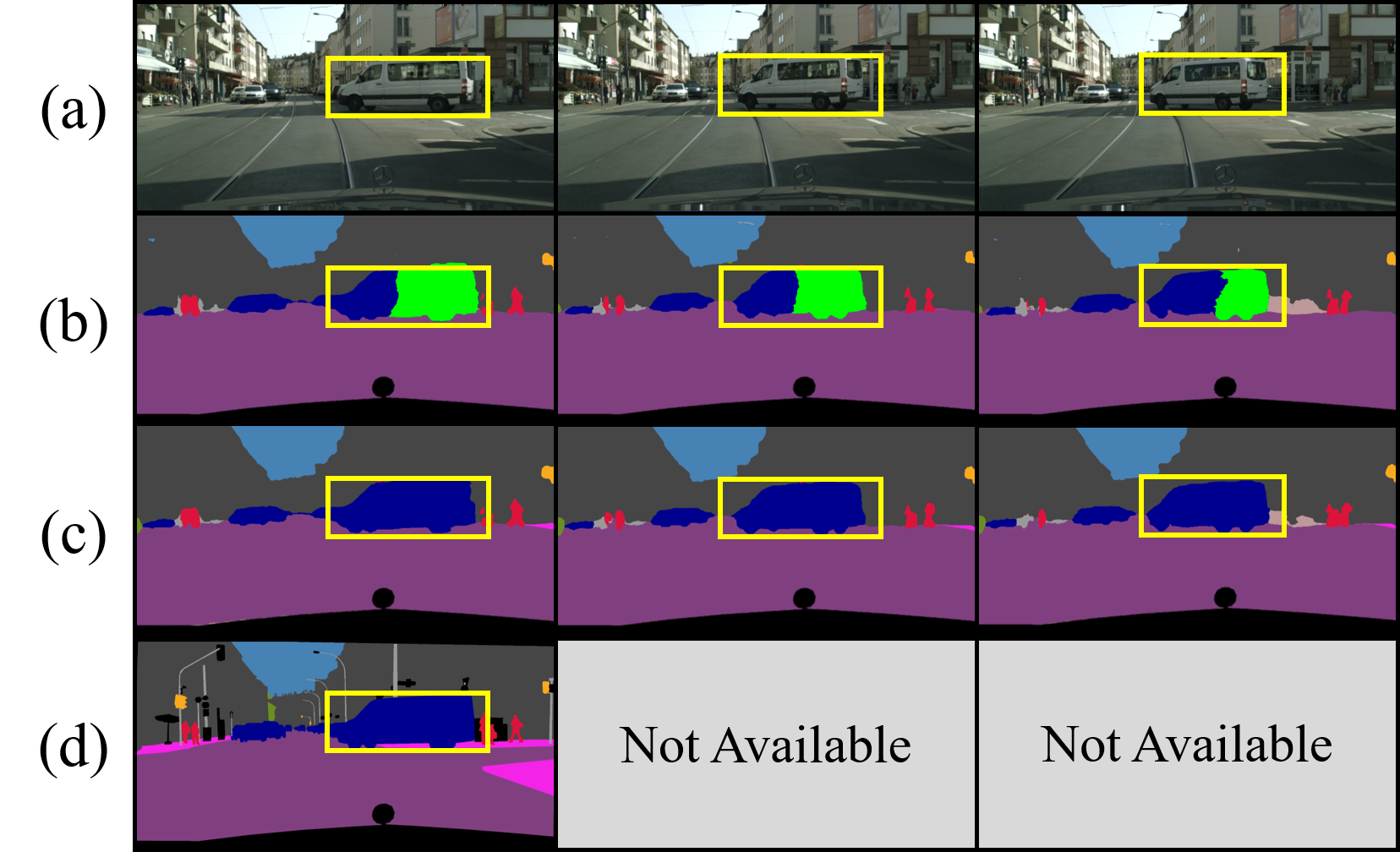}
    \vspace{-5pt}
    \caption{Qualitative comparison on cross-dataset setting, VSPW~\cite{vspw} $\rightarrow$ Cityscape~\cite{cityscapes}. (a) Frames, (b) ISS baseline, (c) DiTTA (ours), (d) GT.}
    \label{fig:qual_cityscape}
    \vspace{-16pt}
\end{figure}

\subsection{Cross-dataset Scenario}
To assess generalizability under domain shift, we evaluate DiTTA in cross-dataset scenario.
Specifically, we adapt an ISS model pre-trained on VSPW~\cite{vspw} to the Cityscapes~\cite{cityscapes} validation set, without any retraining on the target domain.

As shown in Table~\ref{tab:quan_city}, DiTTA consistently outperforms existing test-time adaptation method and even the VSS model CFFM, achieving improvements of $+2.7$ mIoU and $+3.7$ wIoU over the ISS baseline. This notable gain highlights DiTTA's strong ability to perserve segmentation consistency evne when exposed to unseen domain. 

As illustrated in Figure~\ref{fig:qual_cityscape}, the ISS baseline often misclassifies different parts of a single object into multiple classes, leading to fragmented predictions and inconsistent boundaries across frames.
In contrast, DiTTA accurately assigns the correct class to the entire object region and preserves temporal consistency across consecutive frames.
This result indicates that DiTTA not only improves per-frame segmentation accuracy but also effectively stabilizes object-level predictions under cross-dataset domain shifts, where appearance and layout distributions differ substantially.

\subsection{Using ISS Model Trained on Non-video Dataset}
One may wonder whether DiTTA’s performance is partly due to the ISS model being pre-trained on VSPW~\cite{vspw}, which is a video dataset.
While the used ISS model does not access any temporal information during training, we want to clarify that DiTTA’s effectiveness does not stem from implicit exposure to video-specific priors.

\begin{figure}[t!]
    \centering
    \includegraphics[width=0.99\linewidth]{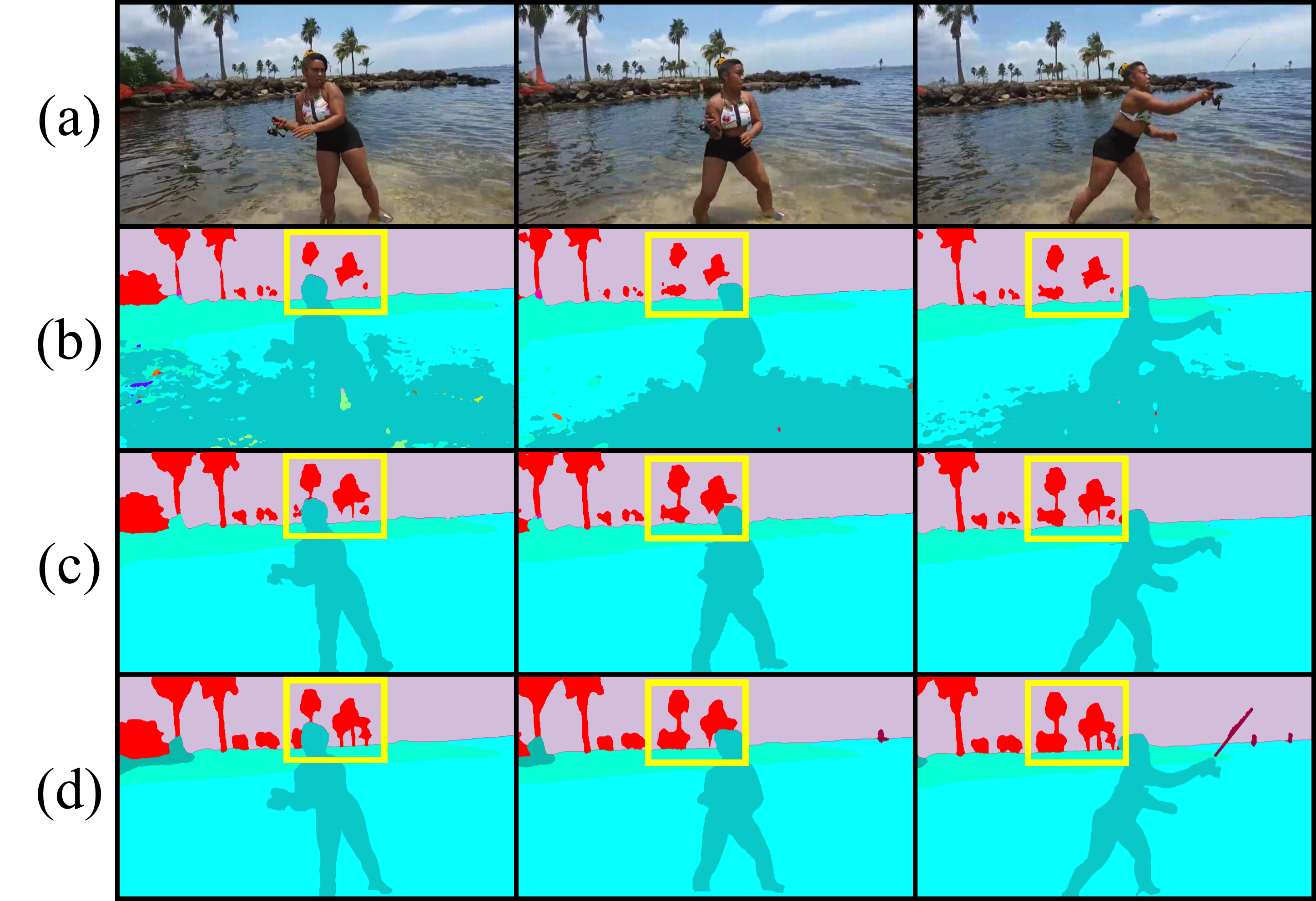}
    \vspace{-5pt}
    \caption{Qualitative comparison on ADE20k~\cite{ade20k} $\rightarrow$ VSPW~\cite{vspw}. (a) Frames, (b) ISS, (c) DiTTA (Ours), (d) GT.}
    \label{fig:qual_ade20k}
    \vspace{-5pt}
\end{figure}

To rule this out, we conduct an additional experiment using an ISS model trained solely on ADE20K~\cite{ade20k}, a purely static image dataset without any temporal structure.
As shown in Table~\ref{tab:ade20k} and Fig.~\ref{fig:qual_ade20k}, DiTTA still achieves strong video segmentation performance on VSPW, despite being initialized from a model trained without any video data.
This result confirms that DiTTA does not rely on hidden temporal bias of video dataset.

\begin{table}[t]
\caption{Experiments with non-video dataset. Methods are trained on ADE20K~\cite{ade20k} and evaluated on VSPW~\cite{vspw}.} 
\centering
\vspace{-6pt}
\resizebox{0.99\linewidth}{!}{
\renewcommand{\tabcolsep}{3pt}
\begin{tabular}{c|cccc}
\hline
\textbf{Methods}  &  \textbf{mIoU} $\uparrow$  &\textbf{ wIoU} $\uparrow$  &  \textbf{mVC$_8$} $\uparrow$  & \textbf{mVC$_{16}$} $\uparrow$  \\ \hline\hline
ISS   & 25.7  & 43.8 & 75.6 &  68.3 \\ \hline
\rowcolor{grey1}DiTTA  & 26.8 \textcolor{blue}{(+1.1)}  & 45.2 \textcolor{blue}{(+1.4)} & 89.0 \textcolor{blue}{(+13.4)} & 86.0 \textcolor{blue}{(+17.7)} \\
\hline
\end{tabular}
}
\label{tab:ade20k}
\vspace{-5pt}
\end{table}

\subsection{Full-video Adaptation}
While our primary evaluation protocol focuses on realistic settings with partial video access (W2F), we also assess DiTTA under a full-video adaptation protocol for completeness.
In this setting, the model is adapted using all frames of the test video and evaluated across the entire video sequence in an offline setting, assuming no major constraints on computational resource.

As shown in Table~\ref{tab:quan_vss}, DiTTA achieves the highest performance across all metrics, significantly outperforming both ISS and VSS baselines.
Notably, it surpasses strong video-based models such as CFFM~\cite{cffm}, THE-Mask~\cite{the_mask_bmvc23}, CFFM++~\cite{cffmpp} and TV3S~\cite{hesham2025exploiting} as well as the zero-shot refinement using SAM2.
These results confirm that DiTTA is not only practical for constrained, real-time scenarios (W2F), but also scalable to more generous conditions where full-video access is available.
This flexibility allows DiTTA to serve a wide range of deployment settings, from lightweight edge inference to offline batch processing.

\subsection{Experiments Using Other ISS Model}
To verify the generalizability of our method across different architectures, we further evaluate DiTTA on diverse use cases.
Specifically, we conduct an additional experiment using OCRNet~\cite{ocrnet}, in addition to the default SegFormer~\cite{segformer} used in the main experiments.
Table~\ref{tab:ocrnet_exp} shows the quantitative performance results of the ISS baseline and DiTTA under the 50\% ratio W2F protocol and full video adaptation.
We utilize the same hyper-parameter values from the experiment using SegFormer, as described in the main paper, in the OCRNet experiment as well.
In both settings, DiTTA demonstrates a consistent performance increase across all metrics.
These results highlight the model-agnostic property of DiTTA, showcasing its practical applicability.

\begin{table}[t!]
\centering
\caption{Experiments for full-video adaptation setting (VSPW \textit{val} set). * denotes re-implemented results.}
\vspace{-7pt}
\label{tab:quan_vss}
\resizebox{0.99\linewidth}{!}{
\renewcommand{\tabcolsep}{2pt}
\begin{tabular}{c|c|c|c|c|c}
\hline
\textbf{Approach}& \textbf{Methods} & \textbf{mIoU} $\uparrow$ & \textbf{wIoU} $\uparrow$ & \textbf{mVC$_8$} $\uparrow$ & \textbf{mVC$_{16}$} $\uparrow$ \\
\hline \hline
ISS & SegFormer~\cite{segformer} & 49.0 & 66.3 & 88.2 & 84.2 \\
\hline
\multirow{4}{*}{VSS}
& CFFM~\cite{cffm} & 49.3 & 65.8 & 90.8 & 87.1 \\
& THE-Mask~\cite{the_mask_bmvc23} & 52.1 & 67.2 & - & - \\
& CFFM++~\cite{cffmpp} & 50.1 & 66.5 & 90.8 & 87.4 \\
& TV3S~\cite{hesham2025exploiting} & 49.8 & - & 91.7 & 88.7 \\
\hline
\multirow{3}{*}{ISS+TTA}
 & TENT*~\cite{wang2020tent} & 49.3 & 65.9 & 87.4 & 83.3 \\
 & CoTTA*~\cite{cotta} & 49.4 & 65.9 & 87.4 & 83.3 \\
& AuxAdapt*~\cite{auxadapt_cvpr} & 50.0 & 66.8 & 92.4 & 89.2 \\
\hline
\multirow{2}{*}{ISS + SAM2}      & Zero-shot Refine. & 49.2 & 66.5 & 94.7 & 93.0 \\
&\cellcolor{grey1} DiTTA (Ours) & \cellcolor{grey1}53.2 & \cellcolor{grey1}68.1 & \cellcolor{grey1}95.9& \cellcolor{grey1}94.3  \\
\hline
\end{tabular}
\vspace{-20pt}
}
\end{table}

\begin{table}[t!]
\centering
\caption{Experiments using OCRNet~\cite{ocrnet} as ISS baseline (VSPW \textit{val} set under 50\% W2F protocol and full video).}
\vspace{-5pt}
\label{tab:ocrnet_exp}
\resizebox{0.99\linewidth}{!}{
\renewcommand{\tabcolsep}{4pt}
\begin{tabular}{c|c|c|c|c|c}
\hline
\textbf{Setting}& \textbf{Aproach} & \textbf{mIoU} $\uparrow$ & \textbf{wIoU} $\uparrow$ & \textbf{mVC$_8$} $\uparrow$ & \textbf{mVC$_{16}$} $\uparrow$ \\
\hline \hline
\multirow{2}{*}{50\%}& OCRNet~\cite{ocrnet} & 34.5 & 56.8 & 81.5 & 73.6 \\
& \cellcolor{grey1}DiTTA (Ours) & \cellcolor{grey1}36.1 & \cellcolor{grey1}58.5 & \cellcolor{grey1}89.2 & \cellcolor{grey1}83.4 \\
\hline
\multirow{2}{*}{Full}       & OCRNet~\cite{ocrnet} & 34.9 & 57.4 & 82.0 & 76.0 \\
&\cellcolor{grey1} DiTTA (Ours) & \cellcolor{grey1}38.1 & \cellcolor{grey1}59.5 & \cellcolor{grey1}89.3 & \cellcolor{grey1}83.5  \\
\hline
\end{tabular}
}
\vspace{-12pt}
\end{table}
\section{Conclusion}
\label{sec:con}
We propose \textbf{DiTTA}, a novel framework that transforms an ISS model into a temporally coherent VSS model by leveraging TTA from SAM2.
Through a single adaptation phase, DiTTA effectively internalizes temporal consistency without requiring any labeled video data or architectural retraining.
DiTTA demonstrates strong capability even in highly data-limited adaptation settings, consistently outperforming both ISS and fully supervised VSS models across various scenarios.
These results establish DiTTA as a data-efficient, annotation-free, and model-agnostic alternative for achieving high-quality video segmentation from image models.

\section*{Acknowledgments}
This work was supported by the Technology Innovation Program (2410013617, 20024355, Development of autonomous driving connectivity technology based on sensor-infrastructure cooperation) funded by the Ministry of Trade, Industry \& Energy(MOTIE, Korea) and the National Research Foundation of Korea(NRF) grant funded by the Korea government(MSIT) (RS-2026-25478915).

{
    \small
    \bibliographystyle{ieeenat_fullname}
    \bibliography{main}
}


\end{document}